\documentclass[11pt,a4paper]{article}
\usepackage{times,latexsym}
\usepackage{url}
\usepackage[T1]{fontenc}
\usepackage{dsfont}
\usepackage{mathtools}
\usepackage{booktabs}
\usepackage{comment}
\usepackage{framed}
\usepackage{varwidth}
\usepackage{array}
\usepackage{hyperref}
\usepackage{xcolor}
\usepackage{amsmath,amsfonts,amssymb,verbatim,ifthen,amsthm}
\usepackage{color}
\usepackage{multirow}
\usepackage[all]{xy}
\usepackage{setspace}
\usepackage{graphicx}
\usepackage{nameref}
\usepackage{bbm}
\usepackage{csquotes}
\usepackage[leftmargin=3em]{quoting}
\usepackage{subcaption}
\usepackage{placeins}
\usepackage{lscape} 
\usepackage{enumitem}
\usepackage{arydshln}
\usepackage{tabularx}
\usepackage{multirow}
\usepackage{xcolor}
\usepackage{comment}
\usepackage{notation}
\usepackage{tikz}
\usepackage{natbib}
\usepackage{authblk}
\usepackage[nameinlink]{cleveref}
\usetikzlibrary{patterns, shapes.geometric, positioning, bayesnet}

\usepackage[acceptedWithA]{tacl2021v1}
\usepackage{mathbbol}
\usepackage[normalem]{ulem}

\DeclareRobustCommand{\parhead}[1]{\textbf{#1}~}

\newtheorem{example}{Example}
\newcommand{\com}[1]{}
\crefname{section}{\S}{\S\S}
\crefname{subsection}{\S}{\S\S}
\crefname{subsubsection}{\S}{\S\S}

\begin{document}

\title{Causal Inference in Natural Language Processing: \\ 
Estimation, Prediction, Interpretation and Beyond}

\author[1,10]{Amir Feder\thanks{All authors equally contributed to this paper. Author names are organized alphabetically in two clusters: First students and post-docs and then faculty members. The email address of the corresponding (first) author is: \texttt{feder@campus.technion.ac.il.}}}
\author[2]{Katherine A. Keith}
\author[3]{Emaad Manzoor}
\author[4]{Reid Pryzant}
\author[5]{Dhanya Sridhar} 
\author[6]{\\ Zach Wood-Doughty}
\author[7]{Jacob Eisenstein}
\author[8]{Justin Grimmer}
\author[1]{Roi Reichart}
\author[9]{Margaret E. Roberts}
\author[10]{\\ Brandon M. Stewart}
\author[7,11]{Victor Veitch}
\author[12]{Diyi Yang}

\affil[1]{Technion - Israel Institute of Technology}
\affil[2]{Williams College}
\affil[3]{University of Wisconsin - Madison}
\affil[4]{Microsoft}
\affil[5]{Columbia University}
\affil[6]{Northwestern University}
\affil[7]{Google Research}
\affil[8]{Stanford University}
\affil[9]{University of California San Diego}
\affil[10]{Princeton University}
\affil[11]{University of Chicago}
\affil[12]{Georgia Tech}

\maketitle

\begin{abstract}
A fundamental goal of scientific research is to learn about causal relationships. However, despite its critical role in the life and social sciences, causality has not had the same importance in Natural Language Processing (NLP), which has traditionally placed more emphasis on predictive tasks. This distinction is beginning to fade, with an emerging area of interdisciplinary research at the convergence of causal inference and language processing. Still, research on causality in NLP remains scattered across domains without unified definitions, benchmark datasets and clear articulations of the challenges and opportunities in the application of causal inference to the textual domain, with its unique properties.
In this survey, we consolidate research across academic areas and situate it in the broader NLP landscape. 
We introduce the statistical challenge of estimating causal effects with text, encompassing settings where text is used as an outcome, treatment, or to address confounding. In addition, we explore potential uses of causal inference to improve the robustness, fairness, and interpretability of NLP models. We thus provide a unified overview of causal inference for the NLP community. \footnote{An online repository containing existing research on causal inference and language processing is available here: \url{https://github.com/causaltext/causal-text-papers}}

\end{abstract}

\section{Introduction}
\label{sec:intro}
The increasing effectiveness of NLP has created exciting new opportunities for interdisciplinary collaborations, bringing NLP techniques to a wide range of external research disciplines~\citep[e.g.,][]{roberts2014structural,zhang2020quantifying,ophir2020deep} and incorporating new data and tasks into mainstream NLP~\citep[e.g.,][]{thomas-etal-2006-get,pryzant2018deconfounded}. 
In such interdisciplinary collaborations, many of the most important research questions relate to the inference of causal relationships. For example, before recommending a new drug therapy, clinicians want to know the causal effect of the drug on disease progression. Causal inference involves a question about a counterfactual world created by taking an intervention: What would a patient's disease progression have been if we had given them the drug? As we explain below, with observational data, the causal effect is not equivalent to the correlation between whether the drug is taken and the observed disease progression. There is now a vast literature on techniques for making valid inferences using traditional (non-text) datasets~\cite[e.g.,][]{morgan2015counterfactuals}, but the application of these techniques to natural language data raises new fundamental challenges. 

Conversely, in many classical NLP applications, the main goal is to make accurate predictions: Any statistical correlation is admissible, regardless of the underlying causal relationship. However, as NLP systems are increasingly deployed in challenging and high-stakes scenarios, we cannot rely on the usual assumption that training and test data are identically distributed, and we may not be satisfied with uninterpretable black-box predictors. For both of these problems, causality offers a promising path forward: Domain knowledge of the causal structure of the data generating process can suggest inductive biases that lead to more robust predictors, and a causal view of the predictor itself can offer new insights on its inner workings. 

The core claim of this survey paper is that deepening the connection between causality and NLP has the potential to advance the goals of both social science and NLP researchers.
We divide the intersection of causality and NLP into two areas: Estimating causal effects from text, and using causal formalisms to make NLP methods more reliable. We next illustrate this distinction.

\begin{example}
\label[example]{eg:1}
An online forum has allowed its users to indicate their preferred gender in their profiles with a female or male icon. They notice that users who label themselves with the female icon tend to receive fewer ``likes'' on their posts. To better evaluate their policy of allowing gender information in profiles, they ask: Does 
using the female icon
cause a decrease in popularity for a post?
\end{example}
Ex.~\ref{eg:1} addresses the causal effect of signaling female gender
 (treatment) on the likes a post receives (outcome) (see discussion on signaling at \cite{keith2020text}). The counterfactual question is: If we could manipulate the gender icon of a post, how many likes would the post have received? 

The observed correlation between the gender icons and the number of ``likes'' generally does not coincide with the causal effect: it might instead be a \emph{spurious correlation}, induced by other variables, known as confounders, which are correlated with both the treatment and the outcome~\citep[see][for an early discussion of spurious correlation in NLP]{gururangan2018annotation}. One possible confounder is the topic of each post: Posts written by users who have selected the female icon may be about certain topics (e.g., child birth or menstruation) more often, and those topics may not receive as many likes from the audience of the broader online platform. As we will see in \Cref{sec:tutorial}, due to confounding, estimating a causal effect requires assumptions.

\Cref{eg:1} highlights the setting where the text encodes the relevant confounders of a causal effect.
The text as a confounder setting is one of many causal inferences we can make with text data. The text data can also encode outcomes or treatments of interest. For example, we may wonder about how gender signal affects the sentiment of the reply that a post receives (text as outcome), or about how a writing style affects the ``likes'' a post receives (text as treatment).

\parhead{NLP helps causal inference.} Causal inference with text data involves several challenges that are distinct from typical causal inference settings: Text is high-dimensional, needs sophisticated modeling to measure semantically meaningful factors like topic, and demands careful thought to formalize the intervention that a causal question corresponds to. The developments in NLP around modeling language, from topic models \cite{blei2003latent} to contextual embeddings (e.g. \cite{devlin2018bert}), offer promising ways to extract the information we need from text to estimate causal effects. However, we need new assumptions to ensure that the use of NLP methods leads to valid causal inferences. We discuss existing research on estimating causal effects from text and emphasize these challenges and opportunities in \Cref{sec:estimation}.

\begin{example}
\label[example]{eg:2}
A medical research center wants to build a classifier to detect clinical diagnoses from the textual narratives of patient medical records. The records are aggregated across multiple hospital sites, which vary both in the frequency of the target clinical condition and the writing style of the narratives. When the classifier is applied to records from sites that were not in the training set, its accuracy decreases. Post-hoc analysis indicates that it puts significant weight on seemingly irrelevant features, such as formatting markers.
\end{example}
Like Ex. \ref{eg:1}, Ex. \ref{eg:2} also involves a counterfactual question: Does the classifier's prediction change if we intervene to change the hospital site, while holding the true clinical status fixed?
We want the classifier to rely on phrases that express clinical facts, and not writing style. However, in the training data, the clinical condition and the writing style are spuriously correlated, due to the site acting as a confounding variable. For example, a site might be more likely to encounter the target clinical condition due to its location or speciality, and that site might also employ distinctive textual features, such as boilerplate text at the beginning of each narrative. In the training set, these features will be predictive of the label, but they are unlikely to be useful in deployment scenarios at new sites. In this example, the hospital site acts like a confounder: It creates a spurious correlation between some features of the text and the prediction target.

\Cref{eg:2} shows how the lack of robustness can make NLP methods less trustworthy. A related problem is that NLP systems are often black boxes, making it hard to understand how human-interpretable features of the text lead to the observed predictions. In this setting, we want to know if some part of the text (e.g., some sequence of tokens) causes the output of an NLP method (e.g., classification prediction).

\parhead{Causal models can help NLP.} To address the robustness and interpretability challenges posed by NLP methods, we need new criteria to learn models that go beyond exploiting correlations. For example, we want predictors that are invariant to certain changes that we make to text, such as changing the format while holding fixed the ground truth label. There is considerable promise in using causality to develop new criteria in service of building robust and interpretable NLP methods. In contrast to the well-studied area of causal inference with text, this area of causality and NLP research is less well-understood, though well-motivated by recent empirical successes. In \S\ref{sec:prediction}, we cover the existing research and review the challenges and opportunities around using causality to improve NLP.

This position paper follows a small body of surveys that review the role of text data within causal inference \citep{egami2018make,keith2020text}. We take a broader view, separating the intersection of causality and NLP into two distinct lines of research on estimating causal effects in which text is at least one causal variable (\S\ref{sec:estimation}) and using causal formalisms to improve robustness and interpretability in NLP methods (\S\ref{sec:prediction}). After reading this paper, we envision that the reader will have a broad understanding of: Different types of causal queries and the challenges they present; the statistical and causal challenges that are unique to working with text data and NLP methods; and open problems in estimating effects from text and applying causality to improve NLP methods.
\section{Background}
\label{sec:tutorial}
Both focal problems of this survey (causal effect estimation and causal formalisms for robust and explainable prediction) involve causal inference. The key ingredient to causal inference is defining counterfactuals based on an intervention of interest. We will illustrate this idea with the motivating examples from \
\S\ref{sec:intro}.

\Cref{eg:1} involves online forum posts and the number of likes $Y$ that they receive. We use a binary variable $T$ to indicate whether a post uses a ``female icon'' ($T=1$) or a ``male icon'' ($T=0$). We view the post icon $T$ as the ``treatment'' in this example, but do not assume that the treatment is randomly assigned (it may be selected by the posts' authors). The \textit{counterfactual} outcome $Y(1)$ represents the number of likes a post would have received had it used a female icon. The counterfactual outcome $Y(0)$ is defined analogously.

The \emph{fundamental problem of causal inference} \cite{holland1986statistics} is that we can never observe $Y(0)$ and $Y(1)$ simultaneously for any \emph{unit of analysis}, the smallest
unit about which one wants to make counterfactual
inquiries (e.g.~a post in Ex.~\ref{eg:1}). This problem is what makes causal inference harder than statistical inference and impossible without identification assumptions (see \Cref{subsec:assumptions}).

\Cref{eg:2} involves a trained classifier $f(X)$ that takes a textual clinical narrative $X$ as input and outputs a diagnosis prediction. The text $X$ is written based on the physician's diagnosis $Y$, and is also influenced by the writing style used at the hospital $Z$. 
We want to intervene upon the hospital $Z$ while holding the label $Y$ fixed. The counterfactual narrative $X(z)$ is the text we would have observed had we set the hospital to the value $z$ while holding the diagnosis fixed.
The counterfactual prediction 
$f(X(z))$ 
is the output the trained classifier would have produced had we given the counterfactual review $X(z)$ as input.

\subsection{Causal Estimands}
An analyst begins by specifying target causal quantities of interest, called causal estimands, which typically involve counterfactuals.
In \Cref{eg:1}, one possible causal estimand is the average treatment effect (ATE) \cite{rubin1974estimating},
\begin{align}
\label{eq:ate}
    \textrm{ATE} = \mathbb{E}[Y(1) - Y(0)].
\end{align}
where the expectation is over the generative distribution of posts. The ATE can be interpreted as the change in the number of likes a post would have received, on average, had the post used a female icon instead of a male icon.

Another possible causal effect of interest is the conditional average treatment effect (CATE) \cite{imbens2015causal},
\begin{align}
    \label{eq:cate}
    \textrm{CATE} = \mathbb{E}[Y(1) - Y(0)~ | ~G].
\end{align}
where $G$ is a predefined subgroup of the population. For example, $G$ could be all posts on political topics. In this case, the CATE can be interpreted as the change in the number of likes a post on a political topic would have received, on average, had the post used a male icon instead of a female icon. CATEs are used to quantify the heterogeneity of causal effects in different population subgroups.

\subsection{Identification Assumptions for Causal Inference}
\label{subsec:assumptions}

We will focus on \Cref{eg:1} and the ATE in \Cref{eq:ate} to explain the assumptions needed for causal inference. Although we focus on the ATE, related assumptions are needed in some form for all causal estimands. Variables are the same as those defined previously in this section.

\parhead{Ignorability} requires that the treatment assignment be statistically independent of the counterfactual outcomes,
\begin{align}
T \indep Y(a) \quad \forall a \in \{0,1\}.
\end{align}
Note that this assumption is not equivalent to independence between the treatment assignment and the \textit{observed} outcome $Y$. For example, if ignorability holds, $Y \indep T$ would \textit{additionally} imply that the treatment has no effect.

Randomized treatment assignment guarantees ignorability by design. For example, we can guarantee ignorability  in \Cref{eg:1} by flipping a coin to select the icon for each post, and disallowing post authors from changing it.

Without randomized treatment assignment, ignorability could be violated by \textit{confounders}, variables that influence both the treatment status and potential outcomes. In \Cref{eg:1}, suppose that: (i) the default post icon is male, (ii) only experienced users change the icon for their posts based on their gender, (iii) experienced users write posts that receive relatively more likes. In this scenario, the experience of post authors is a confounder: posts having female icons are more likely to be written by experienced users and thus, receive more likes. In the presence of confounders, causal inference is only possible if we assume \textit{conditional ignorability},
\begin{equation} \label{eq:cond-ignore}
T \indep Y(a) ~|~ X \quad\forall a \in \{0,1\}
\end{equation}
where $X$ is a set of observed variables, conditioning on which ensures independence between the treatment assignment and the potential outcomes. In other words, we can assume that all confounders are observed.

\parhead{Positivity} requires that the probability of receiving treatment is bounded away from 0 and 1 for all values of the confounders $X$: 
\begin{align}
0 < \text{Pr}(T=1 \mid X=x) < 1, \forall x
\end{align}
Intuitively, positivity requires that each unit under study has the possibility of being treated and has the possibility of being untreated.
Randomized treatment assignment can also guarantee positivity by design.

\parhead{Consistency} requires that the outcome observed for each unit under study at treatment level $a \in \{0,1\}$ is identical to the outcome we would have observed had that unit been assigned to treatment level $a$,
\begin{align} \label{eq:consistency}
T = a \Leftrightarrow Y(a) = Y \quad \forall a \in \{0,1\}
\end{align}
Consistency ensures that the potential outcomes for each unit under study take on a single value at each treatment level.
Consistency will be violated if different unobservable ``versions'' of the treatment lead to different potential outcomes. For example, if red and blue female icons had different effects on the number of likes received, but icon color was not recorded.
Consistency will also be violated if the treatment assignment of one unit affects the potential outcomes of another; a phenomenon called \textit{interference} \cite{rosenbaum2007interference}.
Randomized treatment assignment does not guarantee consistency by design.
e.g. if different icon colors affect the number of likes but are not considered by the model, then a randomized experiment will not solve the problem.
As \citet{hernan2016does} discusses, consistency assumptions are a ``matter of expert agreement'' and, while subjective, these assumptions are at least made more transparent by causal formalisms.

These three assumptions enable identifying the ATE defined in \Cref{eq:ate}, as formalized in the following identification proof:
\begin{align}
    \mathbb{E}[Y(a)]\quad\nonumber
    &\stackrel{(i)}{=}
    \mathbb{E}_X[\mathbb{E}[Y(a)~|~X]]\nonumber \\
    &\stackrel{(ii)}{=}
    \mathbb{E}_X[\mathbb{E}[Y(a)~|~X, T=a]]\nonumber\\
    &\stackrel{(iii)}{=}
    \mathbb{E}_X[\mathbb{E}[Y~|~X, T=a]], \forall a \in \{0,1\}\nonumber
   \label{eq:identification}
\end{align}
where equality (i) is due to iterated expectation, equality (ii) follows from conditional ignorability, and equality (iii) follows from consistency and positivity, which ensures that the conditional expectation $\mathbb{E}[Y~ |~X,~T=a]$ is well defined. 
The final expression can be computed from observable quantities alone.

We refer to other background material to discuss how to identify and estimate causal effects with these assumptions in hand \citep{rubin2005causal,pearl2009causality,imbens2015causal,egami2018make,keith2020text}.

\subsection{Causal Graphical Models}

Finding a set of variables $X$ that ensure conditional ignorability is challenging, and requires making several carefully assessed assumptions about the causal relationships in the domain under study. Causal directed-acyclic graphs (DAGs) \cite{pearl2009causality} enable formally encoding these assumptions and deriving the set of variables $X$ after conditioning on which ignorability is satisfied.

In a causal DAG, an edge $X \rightarrow Y$ implies that $X$ \emph{may or may not} cause $Y$. 
The absence of an edge between $X$ and $Y$ implies that $X$ \emph{does not} cause $Y$. Bi-directed dotted arrows between variables indicate that they are correlated  potentially through some unobserved variable. 

\begin{figure}
    \centering
    \includegraphics[scale=0.47]{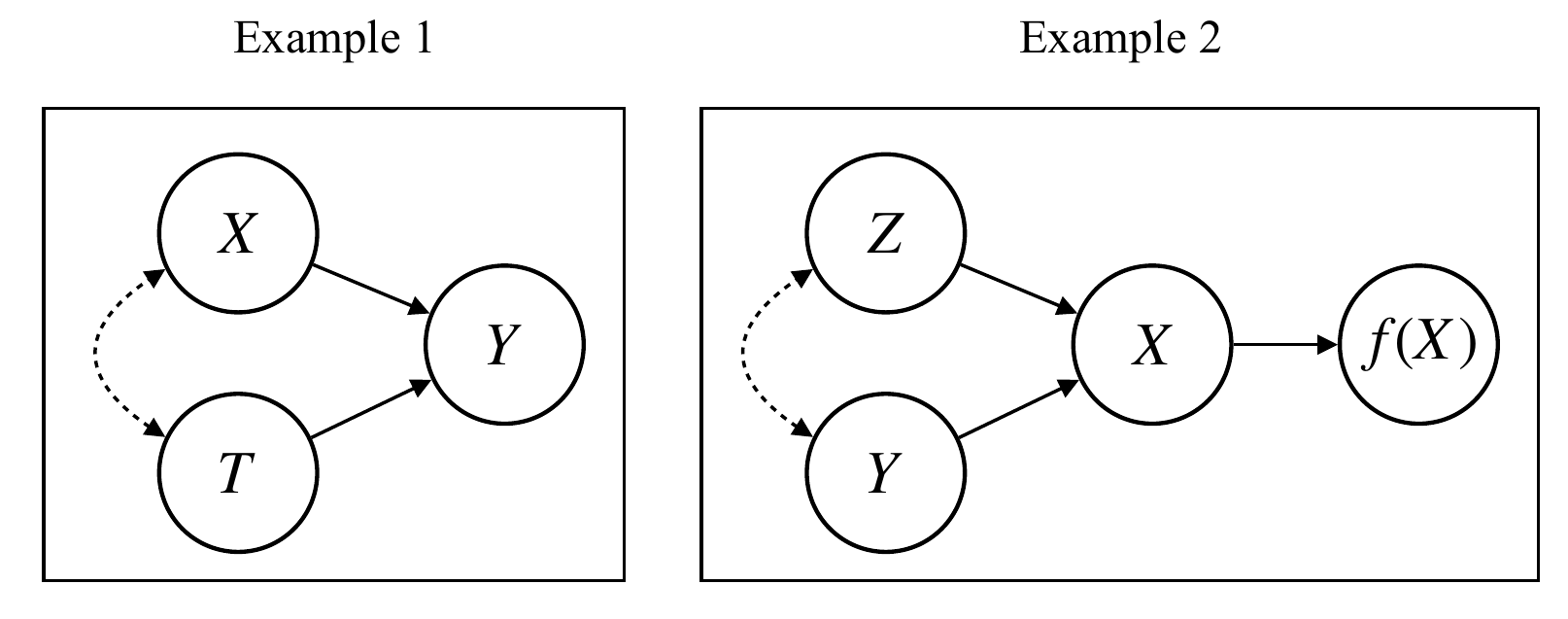}
    \caption{Causal graphs for the motivating examples. (\textbf{Left}) In \Cref{eg:1}, the 
    post icon ($T$) is correlated with attributes of the post ($X$), and both variables affect the number of likes a post receives ($Y$). (\textbf{Right}) 
    In \Cref{eg:2}, the label ($Y$, i.e., diagnosis) and hospital site ($Z$) are correlated, and both affect the clinical narrative ($X$). Predictions $f(X)$ from the trained classifier depend on $X$. 
    \label{fig:causal_dags}}
\end{figure}

\Cref{fig:causal_dags} illustrates the causal DAGs we assume for \Cref{eg:1} and \Cref{eg:2}.
Given a causal DAG, causal dependencies between any pair of variables can be derived using the d-separation algorithm \citep{pearl1994probabilistic}. These dependencies can then be used to assess whether conditional ignorability holds for a given treatment, outcome, and set of conditioning variables $X$. For example, in the left DAG in \Cref{fig:causal_dags}, the post icon $T$ is not independent of the number of likes $Y$ unless we condition on $X$. In the right DAG, the prediction $f(X)$ is not independent of the hospital $Z$ even after conditioning on the narrative $X$. 

\section{Estimating Causal Effects with Text} 
\label{sec:estimation}

In \S\ref{sec:tutorial}, we described assumptions for causal inference when the treatment, outcome and
confounders were directly measured.
In this section, we contribute a novel discussion about how causal assumptions
are complicated when variables necessary for a causal analysis are extracted
automatically from text. Addressing these open challenges will require
collaborations between the NLP and causal estimation communities to understand
what are the requisite assumptions to draw valid causal conclusions. We highlight
prior approaches and future challenges in settings where the text is a
confounder, the outcome, or the treatment -- but this discussion applies
broadly to many text-based causal problems. 

To make these challenges clear, we will expand upon \Cref{eg:1} by supposing
a hypothetical online forum wants to understand and reduce harassment on its
platform.  Many such questions are causal: Do gendered icons influence the
harassment users receive?  Do longer suspensions make users less likely to
harass others? How can a post be rewritten to avoid offending others?
In each case, using NLP to measure aspects of language is integral to any
causal analysis.

\subsection{Causal Effects with Textual Confounders}

Returning to \Cref{eg:1}, suppose the platform worries that users with
female icons are more likely to receive harassment from other users. Such a
finding might significantly influence plans for a new moderation strategy \citep{jhaver2018online,rubin2020fragile}.
We may be unable or unwilling to randomize our treatment (the gender signal) of the author's icon), so the causal effect of gender signal on harassment received might be confounded by other variables. The topic of the post may be
an important confounder: some subject areas may be discussed by a larger
proportion of users with female icons, and more controversial subjects may
attract more harassment. The text of the post provides evidence of the topic
and thus acts as a confounder \citep{roberts2020adjusting}.

\paragraph{Previous approaches.}
The main idea in this setting is to use NLP
methods to extract confounding aspects from text and then adjust
for those aspects in an estimation approach such as propensity score matching.
However, how and when these methods violate causal assumptions are still open questions.
\citet{keith2020text} provides a recent overview of several such methods
and many potential threats to inference.

One set of methods apply unsupervised dimensionality reduction methods
that reduce high-dimensional text data to a low-dimensional set of variables.
Such methods include latent variable models such as topic models, embedding
methods, and auto-encoders. \citet{roberts2020adjusting} and
\citet{sridhar2019estimating} have applied topic models to extract confounding
patterns from text data, and performed an adjustment for these inferred
variables. \citet{mozer2020matching} matches texts using distance metrics on the
bag-of-words representation.

A second set of methods adjust for confounders from text with supervised NLP methods. Recently, \citet{veitch2020using} adapted pre-trained language models and supervised topic models with multiple classification heads for binary treatment and counterfactual outcomes. By learning a ``sufficient'' embedding that obtained low classification loss on the treatment and counterfactual outcomes, they show that confounding properties could be found within text data.  \citet{roberts2020adjusting} combines these strategies with the topic model approach in a text matching framework.

\paragraph{Challenges for causal assumptions with text.}
In settings
without randomized treatments, NLP methods that adjust for text confounding
require a particularly strong statement of conditional ignorability
(Equation~\ref{eq:cond-ignore}): all aspects of confounding must be measured by
the model. Because we cannot test this assumption, we should seek domain
expertise to justify it or understand the theoretical and empirical
consequences if it is violated.  

When the text is a confounder, its
high-dimensionality makes positivity unlikely to hold \citep{d2020overlap}.
Even for approaches that extract a low-dimensional representation of the confounder from text, positivity
is a concern.
For example, in \Cref{eg:1}, posts might contain phrases that near-perfectly encode the chosen gender-icon of the author. If the learned representation captures this information alongside other confounding aspects, it would be nearly impossible to imagine changing the gender icon while holding the gendered text fixed. 

\subsection{Causal Effects on Textual Outcomes}

Suppose platform moderators can choose to suspend users who violate community
guidelines for either one day or one week, and we want to know which option has
the greatest effect at decreasing the toxicity of the suspended user. If we
could collect them for each user's post, ground-truth human annotations of
toxicity would be our ideal outcome variable.  We would then use those outcomes
to calculate the ATE, following the discussion in \Cref{sec:tutorial}.
Our analysis of suspensions is complicated if, instead of ground-truth labels
for our toxicity outcome, we rely on NLP methods to extract the outcome from
the text. A core challenge is to distill the high-dimensional text into a
low-dimensional measure of toxicity. 

\paragraph{Challenges for causal assumptions with text.}
We saw in \Cref{sec:tutorial} that
randomizing the treatment assignment can ensure ignorability and positivity; but
even with randomization, we require more careful assessment to satisfy
consistency. Suppose we randomly assign suspension lengths to users and then
once those users return and continue to post, we use a clustering method to
discover toxic and non-toxic groupings among the formerly-suspended users. To
estimate the causal effect of suspension length, we rely on the trained
clustering model to infer our outcome variable. Assuming that the suspension
policy does in truth have a causal effect on posting behavior, then because our
clustering model depends on all posts in its training data, it also depends on
the treatment assignments that influenced each post. Thus, when we use the
model to infer outcomes, each user's outcome depends on all other users'
treatments. This violates the assumption of consistency -- that potential
outcomes do not depend on the treatment status of other units. This undermines
the theoretical basis for our causal estimate, and in practice, implies that
different randomized treatment assignments could lead to different treatment
effect estimates.

These issues can be addressed by developing the measure on only a
sample of the data and then estimating the effect on a
separate, held-out data sample \cite{egami2018make}.

\subsection{Causal Effects with Textual Treatments}

As a third example, suppose we want to understand what makes a post offensive.
This might allow the platform to provide automated suggestions that encourage
users to rephrase their post. Here, we are interested in the causal effect of
the text itself on whether a reader reports it as offensive. Theoretically, the
counterfactual $Y(t)$ is defined for any $t$, but could be limited to an
exploration of specific \emph{aspects} of the text. For example, do
second-person pronouns make a post more likely to be reported?

\paragraph{Previous approaches.}
One approach to studying the effects of text involves treatment \emph{discovery}: producing interpretable features of the text---such as latent topics or lexical features like n-grams \cite{pryzant2018deconfounded}---that can be causally linked to outcomes. For example, 
\citet{fong2016discovery} discovered features of candidate biographies that drove voter evaluations, \citet{pryzant2017predicting} discovered writing styles in marketing materials that are influential in increasing sales figures, and \citet{zhang2020quantifying} discovered conversational tendencies that lead to positive mental health counseling sessions. 

Another approach is to estimate the causal effects of specific latent properties that are intervened on during an experiment or extracted from text for observational studies \citep{pryzant2020causal,wood2018challenges}.
For example, \citet{gerber2008social} studied the effect of appealing to civic duty on voter turnout.
In this setting, factors are latent properties of the text for which we need a measurement model.

\paragraph{Challenges for causal assumptions with text.}
Ensuring positivity and consistency remains a challenge in this setting,
but assessing conditional ignorability is particularly tricky.
Suppose the treatment is the use of second-person pronouns, but the
relationship between this treatment and the outcome is confounded by other
properties of the text (e.g., politeness). For conditional ignorability to
hold, we would need to extract from the text and condition on all such
confounders, which requires assuming that we can disentangle the treatment from
many other aspects of the text \citep{pryzant2020causal}.  
Such concerns could
be avoided by randomly assigning texts to readers
\cite{fong2016discovery,FonGri21}, but that may be impractical.
Even if we could randomize the assignment of texts, 
we still have to assume that there is no confounding due
to latent properties of the reader, such as their political ideology or their tastes.

\subsection{Future Work}

We next highlight key challenges and opportunities for NLP researchers to facilitate causal inference from text.

\parhead{Heterogeneous effects.}
Texts are read and interpreted differently by different people;  
NLP researchers have studied this problem in the context of heterogeneous perceptions of annotators \cite{paun2018comparing,pavlick2019inherent}. In the field of causal inference, the idea that different subgroups experience different causal effects is formalized by a heterogeneous treatment effect, and is studied using conditional average treatment effects (\Cref{eq:cate}) for different subgroups. It may also be of interest to discover subgroups where the treatment has a strong effect on an outcome of interest. For example, we may want to identify text features that characterize when a treatment such as a content moderation policy is effective. \citet{wager2018estimation} proposed a flexible approach to estimating heterogeneous effects based on random forests. However, such approaches, which are developed with tabular data in mind, may be computationally infeasible for high-dimensional text data. There is an opportunity to extend NLP methods to discover text features that capture subgroups where the causal effect varies.

\parhead{Representation learning.}
Causal inference from text requires extracting low-dimensional features from text. Depending on the setting, the low-dimensional features are tasked with extracting confounding information, outcomes or treatments. The need to measure latent aspects from text connects to the field of text representation learning \cite{le2014distributed,liu2015representation,liu2018learning}.
The usual objective of text representation learning approaches is to model language. Adapting representation learning for causal inference offers open challenges; for example, we might augment the objective function to ensure that (i) positivity is satisfied, (ii) confounding information is not discarded, or (iii) noisily-measured outcomes or treatments enable accurate causal effect estimates.

\parhead{Benchmarks.}
Benchmark datasets have propelled machine learning forward by creating shared metrics by which predictive models can be evaluated. There are currently no real-world text-based causal estimation benchmarks due to the \emph{fundamental problem of causal inference} that we can never obtain counterfactuals on an individual and observe the true causal effects. However, as \citet{keith2020text} discuss, there has been some progress in evaluating text-based estimation methods on semi-synthetic datasets in which real covariates are used to generate treatment and outcomes, e.g.~\citet{veitch2020using,roberts2020adjusting,pryzant2020causal,feder2021causalm,weld2020adjusting}. \citet{wood2021generating} employed large-scale language models for controlled synthetic generation of text on which causal methods can be evaluated. An open problem is the degree to which methods that perform well on synthetic data generalize to real-world data.

\parhead{Controllable Text generation.}
When running a randomized experiment or generating synthetic data, researchers make decisions using the empirical distribution of the data. If we are studying whether a drug prevents headaches, it would make sense to randomly assign a `reasonable' dose -- one that is large enough to plausibly be effective but not so large as to be toxic. But when the causal question involves natural language, domain knowledge might not provide a small set of `reasonable' texts. Instead, we might turn to controllable text generation to sample texts that fulfill some requirements \cite{kiddon2016globally}. Such methods have a long history in NLP; for example, a conversational agent should be able to answer a user's question while being perceived as polite \cite{niu2018polite}. In our text as treatment example where we want to understand which textual aspects make a text offensive, such methods could enable an experiment allowing us to randomly assign texts that differ on only a specific latent aspect. For example, we could change the style of a text while holding its content fixed \cite{logeswaran2018content}. Recent work has explored text generation from a causal perspective \cite{hu2021causal}, but future work could develop these methods for causal estimation.

\section{Robust and Explainable Predictions from Causality}
\label{sec:prediction}

Thus far we have focused on using NLP tools for estimating causal effects in the presence of text data. In this section, we consider using causal reasoning to help solve traditional NLP tasks such as understanding, manipulating, and generating natural language. 

At a first glance, NLP may appear to have little need for causal ideas. The field has achieved remarkable progress from the use of increasingly high-capacity neural architectures to extract correlations from large-scale datasets~\cite{peters2018deep,devlin2018bert,liu2019roberta}. These architectures make no distinction between causes, effects, and confounders, and they make no attempt to identify causal relationships: A feature may be a powerful predictor even if it has no direct causal relationship with the desired output.

Yet correlational predictive models can be untrustworthy~\cite{jacovi2021formalizing}: They may latch onto spurious correlations (``shortcuts''), leading to errors in out-of-distribution (OOD) settings~\cite[e.g.,][]{mccoy2019right}; they may exhibit unacceptable performance differences across groups of users~\cite[e.g.,][]{zhao-etal-2017-men}; and their behavior may be too inscrutable to incorporate into high-stakes decisions~\cite{guidotti2018survey}. 
Each of these shortcomings can potentially be addressed by the causal perspective: Knowledge of the causal relationship between observations and labels can be used to formalize spurious correlations and mitigate their impact
(\cref{subsec:robust});
causality also provides a language for specifying and reasoning about fairness conditions
(\cref{subsec:bias});
and the task of explaining predictions may be naturally formulated in terms of counterfactuals
(\cref{subsec:exp}). 
The application of causality to these problems is still an active area of research, which we attempt to facilitate by highlighting previously implicit connections among a diverse body of prior work.

\subsection{Learning Robust Predictors}
\label{subsec:robust}
The NLP field has grown increasingly concerned with \emph{spurious correlations}
~\cite[][inter alia]{gururangan2018annotation,mccoy2019right}.
From a causal perspective, spurious correlations arise when two conditions are met. First, there must be some factor(s) $Z$ that are informative (in the training data) about both the features $X$ and label $Y$. Second, $Y$ and $Z$ must be dependent in the training data in a way that is not guaranteed to hold in general. 
A predictor $f: X \to Y$ will learn to use parts of $X$ that carry information about $Z$ (because $Z$ is informative about $Y$), which can lead to errors if the relationship between $Y$ and $Z$ changes when the predictor is deployed.\footnote{From the perspective of earlier work on domain adaptation~\citep{sogaard2013semi}, spurious correlations can be viewed as a special case of a more general phenomenon in which feature-label relationships change across domains. For example, the lexical feature \textit{boring} might have a stronger negative weight in reviews about books than about kitchen appliances, but this is not a spurious correlation because there is a direct causal relationship between this feature and the label. Spurious correlations are a particularly important form of distributional shift in practice because they can lead to inconsistent predictions on pairs of examples that humans view as identical.}

This issue is illustrated by \Cref{eg:2}, where the task is to predict a medical condition from the text of patient records. The training set is drawn from multiple hospitals which vary both in the frequency of the target clinical  condition ($Y$) and  the  writing  style  of the narratives (represented in $X$). A predictor trained on such data will use textual features that carry information about the hospital ($Z$), even when they are useless at predicting the diagnosis \emph{within} any individual hospital. 
Spurious correlations also appear as \emph{artifacts} in 
benchmarks for tasks such as natural language inference, where negation words are correlated with semantic contradictions in crowdsourced training data but not in text that is produced under more natural conditions~\cite{gururangan2018annotation,poliak2018hypothesis}. 

Such observations have led to several proposals for novel evaluation methodologies~\cite{naik-etal-2018-stress,ribeiro-etal-2020-beyond,gardner-etal-2020-evaluating} to ensure that predictors are not ``right for the wrong reasons''. These evaluations generally take two forms: \emph{Invariance tests}, which assess whether predictions are affected by perturbations that are causally unrelated to the label, and \emph{sensitivity tests}, which apply perturbations that should in some sense be the minimal change necessary to flip the true label. Both types of test can be motivated by a causal perspective. The purpose of an invariance test is to determine whether the predictor behaves differently on counterfactual inputs $X(Z=\tilde{z})$, where $Z$  indicates a property
that an analyst believes should be causally irrelevant to $Y$. A model whose predictions are invariant across such counterfactuals can in some cases be expected to perform better on test distributions with a different relationship between $Y$ and $Z$~\cite{veitch2021counterfactual}. Similarly, sensitivity tests can be viewed as evaluations of counterfactuals $X(Y=\tilde{y})$, in which the label $Y$ is changed but all other causal influences on $X$ are held constant~\cite{kaushik2020explaining}.
Features that are spuriously correlated with $Y$ will be identical in the factual $X$ and the counterfactual $X(Y=\tilde{y})$. A predictor that relies solely on such spurious correlations will be unable to correctly label both factual and counterfactual instances.

A number of approaches have been proposed for learning predictors that pass tests of sensitivity and invariance. Many of these approaches are either explicitly or implicitly motivated by a causal perspective. They can be viewed as ways to incorporate knowledge of the causal structure of the data into the learning objective.

\subsubsection{Data augmentation}
\label{subsubsec:data}
To learn predictors that pass tests of invariance and sensitivity, a popular and straightforward approach is \emph{data augmentation}: Elicit or construct counterfactual instances, and incorporate them into the training data. When the counterfactuals involve perturbations to confounding factors $Z$, it can help to add a term to the learning objective to explicitly penalize disagreements in the predictions for counterfactual pairs, e.g., ${|f(X(Z=z)) - f(X(Z=\tilde{z}))|}$, when $f$ is the prediction function~\cite{garg2019counterfactual}. When perturbations are applied to the label $Y$, training on label counterfactuals $X(Y=\tilde{y})$ can improve OOD generalization and reduce noise sensitivity  ~\cite{kaushik2019learning,kaushik2020explaining, jha2020does}.\footnote{More broadly, there is a long history of methods that elicit or construct new examples and labels with the goal of improving generalization, e.g. self-training~\cite{mcclosky2006effective,reichart2007self}, co-training~\cite{steedman2003bootstrapping}, and adversarial perturbations~\cite{ebrahimi2018hotflip}. 
The connection of such methods to causal issues such as spurious correlations has not been explored until recently~\cite{chen2020self,jin2021causal}.}

Counterfactual examples can be generated in several ways: (1) manual post-editing~\cite[e.g.,][]{kaushik2019learning,gardner-etal-2020-evaluating}, (2) heuristic replacement of keywords~\cite[e.g.,][]{shekhar-etal-2017-foil,garg2019counterfactual,feder2021causalm}, and (3) automated text rewriting~\cite[e.g.,][]{zmigrod-etal-2019-counterfactual,riley2020textsettr,wu2021polyjuice,Calderon:22}. Manual editing is typically fluent and accurate but relatively expensive. Keyword-based approaches are appropriate in some cases --- for example, when counterfactuals can be obtained by making local substitutions of closed-class words like pronouns --- but they cannot guarantee fluency or coverage of all labels and covariates of interest~\cite{antoniak-mimno-2021-bad}, and are difficult to generalize across languages. Fully generative approaches could potentially combine the fluency and coverage of manual editing with the ease of lexical heuristics.

Counterfactual examples are a powerful resource because they directly address the missing data issues that are inherent to causal inference, as described in \cref{sec:tutorial}.
However, in many cases it is difficult for even a fluent human to produce meaningful counterfactuals: Imagine the task of converting a book review into a restaurant review while somehow leaving ``everything else'' constant (as in \citet{Calderon:22}). 
A related concern is lack of precision in specifying the desired impact of the counterfactual. To revise a text from, say, U.S. to U.K. English, it is unambiguous that ``colors'' should be replaced with ``colours'', but should terms like ``congress'' be replaced with analogous concepts like ``parliament''? This depends on whether we view the semantics of the text as a causal descendent of the locale. If such decisions are left to the annotators' intuitions, it is difficult to ascertain what robustness guarantees we can get from counterfactual data augmentation.
Finally, there is  the possibility that counterfactuals will introduce new spurious correlations. For example, when asked to rewrite NLI examples without using negation, annotators (or automated text rewriters) may simply find another shortcut, introducing a new spurious correlation. Keyword substitution approaches may also introduce new spurious correlations if the keyword lexicons are incomplete~\cite{joshi2021investigation}.
Automated methods for conditional text rewriting are generally not based on a formal counterfactual analysis of the data generating process~\citep[cf.][]{pearl2009causality}, which would require modeling the relationships between various causes and consequences of the text. The resulting counterfactual instances may therefore fail to fully account for spurious correlations and may introduce new spurious correlations.

\subsubsection{Distributional Criteria}
\label{subsubsec:distributional}
An alternative to data augmentation is to design new learning algorithms that operate directly on the observed data. 
In the case of invariance tests, one strategy is to derive distributional properties of invariant predictors, and then ensure that these properties are satisfied by the trained model.

Given observations of the potential confounder at training time, the counterfactually-invariant predictor will satisfy an independence criterion that can be derived from the causal structure of the data generating process~\citep{veitch2021counterfactual}.
Returning to \Cref{eg:2}, the desideratum is that the predicted diagnosis $f(X)$ should not be affected by the aspects of the writing style that are associated with the hospital $Z$. 
This can be formalized as \emph{counterfactual invariance} to $Z$: The predictor $f$ should satisfy $f(X(z))=f(X(z'))$ for all $z,z'$. In this case, both $Z$ and $Y$ are causes of the text features $X$.\footnote{This is sometimes called the \emph{anticausal} setting, because the predictor $f: X \to \hat{Y}$ must reverse the causal direction of the data generating process~\citep{scholkopf2012causal}.} Using this observation, it can be shown that any counterfactually invariant predictor will satisfy ${f(X) \indep Z \mid Y}$,  i.e., the prediction $f(X)$ is independent of the covariate $Z$ conditioned on the true label $Y$.
In other cases, such as content moderation, the label is an effect of the text, rather than a cause --- for a detailed discussion of this distinction, see \citet{jin2021causal}. In such cases, it can be shown that a counterfactually-invariant predictor will satisfy $f(X) \indep Z$ (without conditioning on $Y$). In this fashion, knowledge of the true causal structure of the problem can be used to derive observed-data signatures of the counterfactual invariance. Such signatures can be incorporated as regularization terms in the training objective (e.g., using kernel-based measures of statistical dependence). These criteria do not guarantee counterfactual invariance---the implication works in the other direction---but in practice they increase counterfactual invariance and improve performance in out-of-distribution settings without requiring counterfactual examples.

An alternative set of distributional criteria can be derived by viewing the training data as arising from a finite set of \emph{environments}, in which each environment is endowed a unique distribution over causes, but the causal relationship between $X$ and $Y$ is invariant across environments. This view motivates a set of environmental invariance criteria: The predictor should include a representation function that is invariant across environments~\cite{muandet2013domain,peters2016causal}; we should induce a representation such that the same predictor is optimal in every environment~\cite{arjovsky2019invariant}; the predictor should be equally well calibrated across environments~\citep{wald2021calibration}.  Multi-environment training is conceptually similar to domain adaptation~\cite{ben2010theory}, but here the goal is not to learn a predictor for any specific target domain, but rather to learn a predictor that works well across a set of causally-compatible domains, known as \emph{domain generalization}~\cite{ghifary2015domain,gulrajani2020search}.
However, it may be necessary to observe data from a very large number of environments to disentangle the true causal structure~\citep{rosenfeld2021risks}.

Both general approaches require richer training data than in typical supervised learning: Either explicit labels $Z$ for the factors to disentangle from the predictions or access to data gathered from multiple labeled environments. Obtaining such data may be rather challenging, even compared to creating counterfactual instances. Furthermore, the distributional approaches have thus far been applied only to classification problems, while data augmentation can easily be applied to structured outputs such as machine translation. 

\subsection{Fairness and bias}
\label{subsec:bias}
NLP systems inherit and sometimes amplify undesirable biases encoded in text training data~\cite{barocas-hardt-narayanan,blodgett2020language}. Causality can provide a language for specifying desired fairness conditions across demographic attributes like race and gender. Indeed, fairness and bias in predictive models have close connections to causality: \citet{hardt2016equality} argue that a causal analysis is required to determine the fairness properties of an observed distribution of data and predictions; \citet{kilbertus2017avoiding} show that fairness metrics can be motivated by causal interpretations of the data generating process; \citet{kusner2017counterfactual} study ``counterfactually fair'' predictors where, for each individual, predictions are the same for that individual and for a counterfactual version of them created by changing a protected attribute. However, there are important questions about the legitimacy of treating attributes like race as variables subject to intervention~\cite[e.g.,][]{kohler2018eddie,hanna2020towards}, and \citet{kilbertus2017avoiding} propose to focus instead on invariance to observable proxies such as names. 

\paragraph{Fairness with text.} The fundamental connections between causality and unfair bias have been explored mainly in the context of relatively low-dimensional tabular data rather than text. However, there are several applications of the counterfactual data augmentation strategies from \cref{subsubsec:data} in this setting: For example, \citet{garg2019counterfactual} construct counterfactuals by swapping lists of ``identity terms'', with the goal of reducing bias in text classification, and \citet{zhao-etal-2018-gender} swap gender markers such as pronouns and names for coreference resolution. Counterfactual data augmentation has also been applied to reduce bias in pre-trained models~\cite[e.g.,][]{huang2019reducing,maudslay2019s} but the extent to which biases in pre-trained models propagate to downstream applications remains unclear~\cite{goldfarb-tarrant-etal-2021-intrinsic}. 
Fairness applications of the distributional criteria discussed in \cref{subsubsec:distributional} are relatively rare, but \citet{adragna2020fairness} show that invariant risk minimization~\cite{arjovsky2019invariant}
can reduce the use of spurious correlations with race for toxicity detection. 

\subsection{Causal Model Interpretations}
\label{subsec:exp}
Explanations of model predictions can be crucial to help diagnose errors and establish trust with decision makers~\cite{guidotti2018survey, jacovi2020towards}. One prominent approach to generate explanations is to exploit network artifacts, such as attention weights~\cite{bahdanau2014neural}, which are computed on the path to generating a prediction~\cite[e.g.,][]{xu2015show,wang-etal-2016-attention}. Alternatively, there have been attempts to estimate simpler and more interpretable models by using perturbations of test examples or their hidden representations~\cite{ribeiro2016should, lundberg2017unified, kim2018interpretability}. However, both attention and perturbation-based methods have important limitations. Attention-based explanations can be misleading \cite{jain2019attention}, and are generally possible only for individual tokens; they cannot explain predictions in terms of more abstract linguistic concepts. Existing perturbation-based methods often generate implausible counterfactuals and also do not allow for estimating the effect of sentence-level concepts.

Viewed as a causal inference problem, explanation can be performed by comparing predictions for each example and its generated  counterfactual. 
While it is usually not possible to observe counterfactual predictions, here the causal system is the predictor itself. In those cases it may be possible to compute counterfactuals, e.g. by manipulating the activations inside the network ~\cite{vig2020causal, geiger2021causal}. Treatment effects can then be computed by comparing the predictions under the factual and counterfactual conditions.
Such a controlled setting is similar to the randomized experiment described in \cref{sec:tutorial}, where it is possible to compute the difference between an actual text and what the text would have been had a specific concept not existed in it. Indeed, in cases where counterfactual texts can be generated, we can often estimate causal effects on text-based models~\cite{ribeiro-etal-2020-beyond, gardner-etal-2020-evaluating, rosenberg2021vqa, ross2021tailor, meng2022locating, zhang2022can}. However, generating such counterfactuals is challenging (see \cref{subsubsec:data}).

To overcome the counterfactual generation problem, another class of approaches proposes to manipulate the representation of the text and not the text itself~\cite{feder2021causalm, elazar2021amnesic, ravfogel2021counterfactual}. 
\citet{feder2021causalm} compute the counterfactual representation by pre-training an additional instance of the language representation model employed by the classifier, with an adversarial component designed to "forget" the concept of choice, while controlling for confounding concepts. \citet{ravfogel2020null} offer a method for removing information from representations by iteratively training linear classifiers and projecting the representations on their null-spaces, but do not account for confounding concepts.

A complementary approach is to \emph{generate} counterfactuals with minimal changes that obtain a different model prediction~\cite{wachter2017counterfactual, mothilal2020explaining}. Such examples allow us to observe the changes required to change a model's prediction. Causal modeling can facilitate this by making it possible to reason about the causal relationships between observed features, thus identifying minimal actions which might have downstream effects on several features, ultimately resulting in a new prediction~\citep{karimi2021algorithmic}.

Finally, a causal perspective on attention-based explanations is to view internal nodes as \emph{mediators} of the causal effect from the input to the output~\cite{vig2020causal,finlayson-etal-2021-causal}. By querying models using manually-crafted counterfactuals, we can observe how information flows, and identify where in the model it is encoded. 

\subsection{Future work}
\label{subsec:future}
In general we cannot expect to have full causal models of text, so a critical question for future work is how to safely use \emph{partial} causal models, which omit some causal variables and do not completely specify the causal relationships within the text itself. A particular concern is unobserved confounding between the variables that are explicitly specified in the causal model. Unobserved confounding is challenging for causal inference in general, but it is likely to be ubiquitous in language applications, in which the text arises from the author's intention to express a structured arrangement of semantic concepts, and the label corresponds to a query, either directly on the intended semantics or on those understood by the reader.

Partial causal models of text can be ``top down'', in the sense of representing causal relationships between the text and high-level document metadata such as authorship, or ``bottom up'', in the sense of representing local linguistic invariance properties, such as the intuition that a multiword expression like `San Francisco' has a single cause. The methods described here are almost exclusively based on top-down models, but approaches such as perturbing entity spans~\citep[e.g.,][]{longpre2021entity} can be justified by implicit bottom-up causal models. Making these connections more explicit may yield new insights. Future work may also explore hybrid models that connect high-level document metadata with medium-scale spans of text such as sentences or paragraphs.

A related issue is when the true variable of interest is unobserved but we do receive some noisy or coarsened proxy variable. For example, we may wish to enforce invariance to dialect but have access only to geographical information, with which dialect is only approximately correlated. This is an emerging area within the statistical literature~\citep{tchetgen2020introduction}, and despite the clear applicability to NLP, we are aware of no relevant prior work.

Finally, applications of causality to NLP have focused primarily on classification, so it is natural to ask how these approaches might be extended to structured output prediction. This is particularly challenging for distributional criteria like $f(X) \indep Z \mid Y$, because $f(X)$ and $Y$ may now represent sequences of vectors or tokens. In such cases it may be preferable to focus on invariance criteria that apply to the loss distribution or calibration.

\section{Conclusion}
\label{sec:conclusion}

Our main goal in this survey was to collect the various touchpoints of causality and NLP into one space, which we then subdivided into the problems of estimating the magnitude of causal effects. and more traditional NLP tasks. These branches of scientific inquiry share common goals, intuitions, and are beginning to show methodological synergies. In \cref{sec:estimation} we showed how recent advances in NLP modeling can help researchers make causal conclusions with text data and the challenges of this process. In \cref{sec:prediction}, we showed how ideas from causal inference can be used to make NLP models more robust, trustworthy and transparent. We also gather approaches that are implicitly causal and explicitly show their relationship to causal inference. Both of these spaces, especially the use of causal ideas for robust and explainable predictions, remain nascent with a large number of open challenges which we have detailed throughout this paper.

A particular advantage of causal methodology is that it forces practitioners to explicate their assumptions. To improve scientific standards, we believe that the NLP community should be clearer about these assumptions and analyze their data using causal reasoning. This could lead to a better understanding of language and the models we build to process it.

\bibliography{bibliography}
\bibliographystyle{acl_natbib}

\end{document}